\documentclass[review]{elsarticle}
\usepackage{hyperref}
\usepackage{float}
\usepackage{verbatim} 
\usepackage{apalike}
\restylefloat{figure}
\restylefloat{table}
\usepackage{times}
\usepackage{epsfig}\usepackage{amsmath}
\usepackage{amssymb}
\usepackage{booktabs}
\usepackage{subfig}

\biboptions{authoryear}

\begin{document}
\begin{frontmatter}

\title{Feature boosting with efficient attention for scene parsing}

\author[label1]{Vivek Singh}
\ead{vivek.singh@plymouth.ac.uk}

\author[label2]{Shailza Sharma \corref{cor1}}
\ead{s.sharma@leeds.ac.uk}

\author[label3]{Fabio Cuzzolin}
\ead{fabio.cuzzolin@brookes.ac.uk}

\cortext[cor1]{Corresponding author.}
\address[label1]{School of Engineering, Computing and Mathematics, University of Plymouth, UK}
\address[label2]{School of Earth and Environment, University of Leeds, UK}
\address[label3]{School of Engineering, Computing and Mathematics, Oxford Brookes University, UK}

\begin{abstract}
The complexity of scene parsing grows with the number of object and scene classes, which is higher in unrestricted open scenes. The biggest challenge is to model the spatial relation between scene elements while succeeding in identifying objects at smaller scales. This paper presents a novel feature-boosting network that gathers spatial context from multiple levels of feature extraction and computes the attention weights for each level of representation to generate the final class labels. A novel `channel attention module' is designed to compute the attention weights, ensuring that features from the relevant extraction stages are boosted while the others are attenuated. The model also learns spatial context information at low resolution to preserve the abstract spatial relationships among scene elements and reduce computation cost. Spatial attention is subsequently concatenated into a final feature set before applying feature boosting. Low-resolution spatial attention features are trained using an auxiliary task that helps learning a coarse global scene structure. The proposed model outperforms all state-of-the-art models on both the ADE20K and the Cityscapes datasets.
\end{abstract}


\end{frontmatter}

\section{Introduction}
\label{introduction}

\begin{figure}
\centering
\begin{tabular}{ccc}
\includegraphics[width=3.2cm]{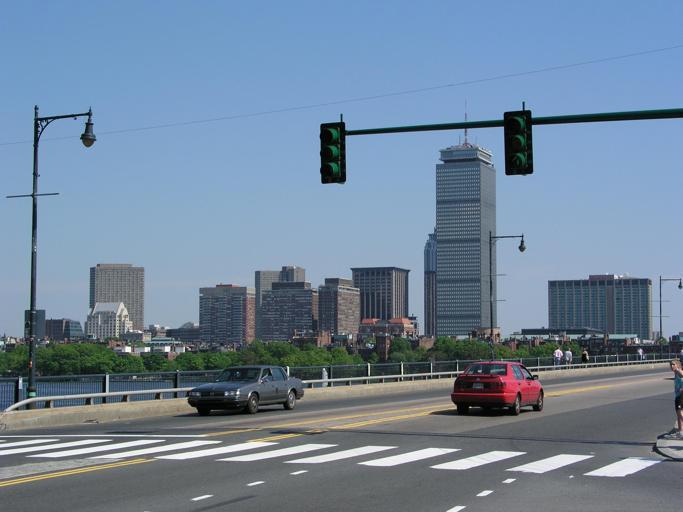}&
\includegraphics[width=3.2cm]{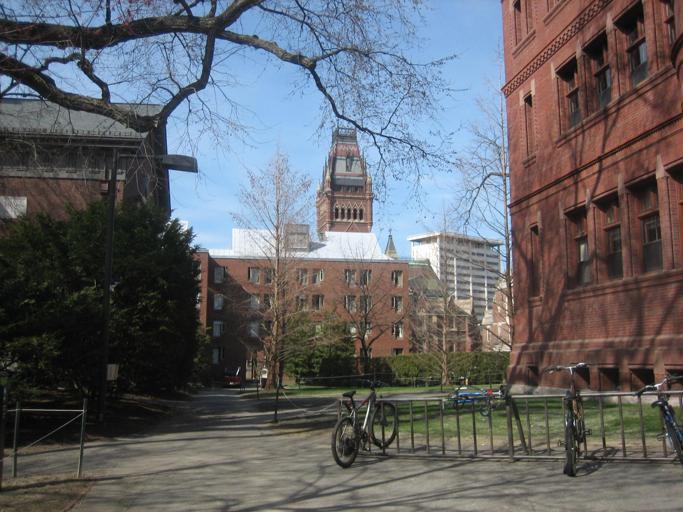}&
\includegraphics[width=3.2cm]{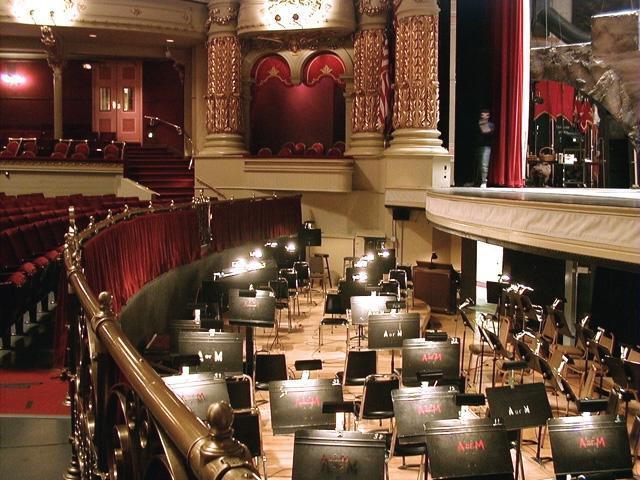}\\
\includegraphics[width=3.2cm]{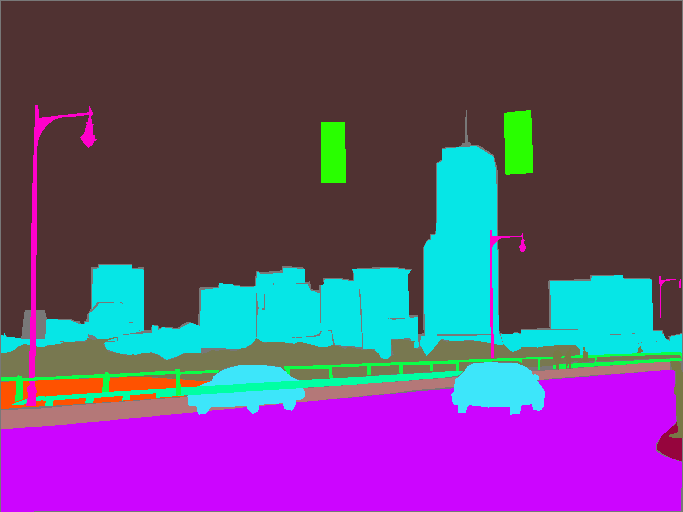}&
\includegraphics[width=3.2cm]{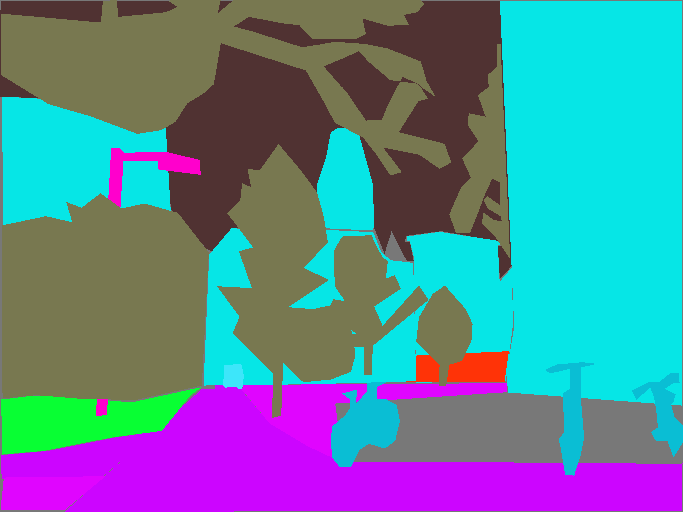}&
\includegraphics[width=3.2cm]{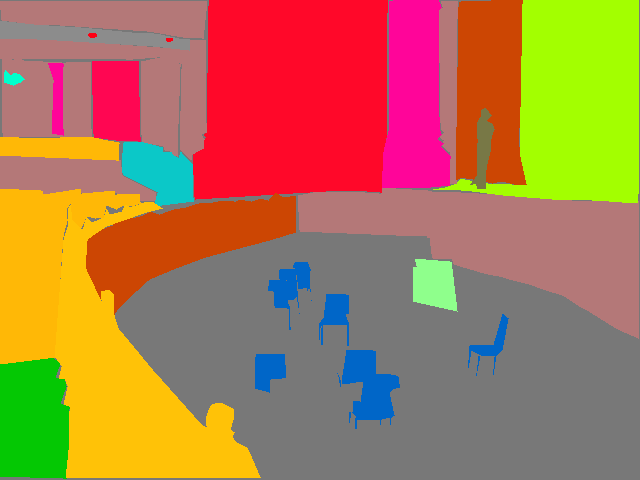}\\
\end{tabular}
    \caption{Samples images from the ADE20K dataset \cite{zhou2017scene} to reflect the complexity of unrestricted natural scenes.}
    \label{fig:ade_sample}
\end{figure}

Scene parsing is the task of breaking an image into its constituent components and is crucial to reach a complete understanding of a visual scene. Recently, semantic image segmentation has gained even more traction within the artificial intelligence community because of the huge scope for a commercial application, e.g., robotic vision, visual scene understanding, aerial image analysis, autonomous driving, etc. \cite{taghanaki2019deep,liu2019recent}. The complexity of the segmentation task, however, can vary as a function of the set of categories being considered for the segmentation and of how fine-grained the description needs to be 
(e.g., exact vehicle model vs generic 'vehicle' class, or breed of dog vs 'dog'). For an input image $I$ of size $H \times W$ the segmentation output will retain the same size (i.e., $H \times W$), but with each location encoding a label (represented by a number from $0$ to $L-1$) instead of color information.

The most recent scene segmentation methods are based on Fully Convolutional Networks (FCNs) \cite{long2015fully} which only use convolutional layers for making the prediction where output contains class-wise probability distribution for each input pixel. The biggest advantage of FCN architectures is that they are scale-independent and computationally very efficient.
FCN-based models have been successful at tackling a number of semantic segmentation task \cite{lin2017refinenet,poudel2019fast,wu2019ace}, especially when number of object classes are limited and scene complexity is manageable. However, as the number of classes increases and unrestricted natural scenes are considered, where dozens of objects are present at various scales and orientations, even state-of-the-art methods struggle to perform reliably. This is the case, for example, with the ADE20K dataset \cite{zhou2017scene}. Fig. \ref{fig:ade_sample} shows some images from ADE20K to illustrate the variability exhibited by natural scenes in both indoor and outdoor settings.

The neighbourhood of each pixel plays a key role in identifying the class label for that pixel.
A key issue with images is that, due to perspective projection, the same object can appear in different sizes or shapes. 
This, in turn, greatly affects the shape and size of the `context window' to be used when considering neighbourhoods.
A fixed-depth network with fixed-size convolutional kernels simply cannot extract 
relationships between pixels when objects can be present at different scale and orientations. 

To address this problem, we propose an architecture that gathers multi-level representations from different stages of feature extraction, providing rich contextual information at different levels of abstraction. The proposed model uses two attention modules: a Channel Attention Module (CAM) and a Spatial Attention Module (SAM). SAM generates spatial attention maps which identify the importance of each feature inside single feature map. Whereas, CAM identifies importance of feature across channels and is applied at the last stage of feature extraction. Before applying channel attention, features from different stages of feature extraction are gathered ($f_i(x,y)$) which provide feature vector ($f_i$) with multi-scale neighbour for each pixel location $(x,y)$.  
Additionally, an auxiliary task is used to train the SAM to learn the low-resolution semantic structure of the scene. The proposed model outperforms all state-of-the-art approaches on the ADE20K \cite{zhou2017scene} and Cityscapes \cite{Cordts2016Cityscapes} benchmark datasets.

In summary, our main contributions are:
\begin{itemize}
    \item 
    A novel feature extraction approach designed to learn and exploit \emph{multi-scale spatial context} for scene parsing.
    \item 
    A novel \emph{channel attention module} is  geared to learn the individual contribution of each feature to the final semantic label. Additionally, a simplified spatial attention module efficiently extracts the relevant attention matrix. 
    \item 
    A new learning mechanism which uses low-resolution semantic maps in an auxiliary task to improve the training of the spatial attention module.
    \item The proposed models shows significantly better performance with much lower parameter count.
\end{itemize}



\vspace{-6mm}

\section{Related work}

The image segmentation literature broadly focuses onto either model architectures or spatial context learning mechanisms. 
The former group of papers focus on techniques for extracting more task-relevant features to improve model generalisation, whereas the latter group considers the relevance of spatial features across adjacent pixels.

\subsection{Segmentation architectures}
The first successful application of Fully Convolutional Neural Network for semantic segmentation is by Lon et al. \cite{long2015fully}. Convolution was used there to extract spatial features and pooling was used to reduce the size of the feature maps. The paper used a deconvolution (up-sampling) operation on the extracted feature to restore model output to the original input's size. 
Subsequent architectures used the encoder-decoder approach proposed in \cite{noh2015learning} for semantic segmentation. In these FCN-based networks 
the encoder block extracts the features from the image while the decoder uses these features to generate a semantic map. 
Architectures such as like U-net \cite{ronneberger2015u}, SegNet \cite{badrinarayanan2017segnet} and V-net \cite{milletari2016v} all use this encoder-decoder concept as the backbone of their designs. Both U-net \cite{ronneberger2015u}
and V-net \cite{milletari2016v} use skip connections between encoder and decoder stages to gather  multi-level feature representation. In SegNet \cite{badrinarayanan2017segnet}, rather than training an upsampling layer, the authors apply an `un-pooling' operation in which feature indexes 
from the max pooling layers are used to upsample the features. 

Pyramid Scene Parsing Networks \cite{zhao2017pyramid} use a Pyramidal Pooling Module (PPM) to learn multi-scale feature representations, idea originally proposed by \cite{long2015fully}. The PPM applies average pooling on the final feature maps with kernels of different sizes. Later outputs of average pooling operations are concatenated together to provide a global prospective. Much recent work uses PPMs as part their architecture
\cite{yu2018semantic,li2018pyramid,yang2018automatic}. The recent DeepLabV3+ approach \cite{chen2018encoder} proposes the use of atrous convolution in the encoder at different dilation rates to generate multi-level context information, resulting in a very effective way of improving model performance. 
The work in \cite{hu2018squeeze} incorporated a `Squeeze and Excite' (SE) block in their architecture to add global context but for classification and detection task. The method re-calibrates the channel-wise feature response of each layer 
by explicitly modeling the interdependence among different channels. RefineNet \cite{lin2017refinenet}, on its side, extracts features at different scales throughout the network's stages and uses this multiscale feature representation to predict the final segmentation labels. 
Another successful architecture, termed UPerNet \cite{xiao2018unified}, uses a unified perceptual parsing approach which tries to learn as many visual concepts as possible from the given data. 
This work leverages multiple dataset such as ADE20k \cite{zhou2017scene}, Pascal-Context \cite{mottaghi2014role}, Pascal-Part \cite{chen2014detect}, OpenSurfaces \cite{bell2013opensurfaces}, and so on, as all these datasets can provide complementary information about the images such as surface texture, objects, or parts. There are other methods that use neural architecture search \cite{zhang2021dcnas} or model adaptation \cite{fleuret2021uncertainty}.


\subsection{Spatial context modeling}
Spatial relationship modeling considers the dependencies among image pixels and how these these dependencies affect the object structure and class label for the given pixel. 
Both Wang et al. \cite{wang2017residual} 
and Zhao et al. \cite{zhao2018psanet} introduced a spatial attention method for image segmentation. The model in \cite{zhao2018psanet}, in particular, uses two attention branches to capture both local and global spatial attention across features. Zhang et al.'s EncNet \cite{zhang2018context}, instead, uses context learning for modeling spatial relationships. The paper also uses an auxiliary task to learn image-level classes 
as, according to authors, this helps improve model performance.
A self attention-based context-learning approach was proposed by \cite{yuan2019object} which led to state-of-the-art performance on the ADE20k dataset \cite{zhou2017scene}. A dual attention architecture was instead proposed in \cite{fu2019dual} which
uses two distinct modules for spatial and channel attention. Both attention modules are applied in parallel on the output of backbone model.  
Attention is applied in \cite{chen2016attention} to the scale of the input, using different levels of scaling on the input image. The scaled versions of the output are then merged together by the attention mechanism. 
Finally, Hu et al. \cite{li2019pancreas} used stack-based attention for CT scan images, in which each slice of the stack gives global contextual information about the ground truth.

The proposed model is novel in both its architecture as well context modeling. The architecture is designed with memory requirements in mind (as shown in Figure \ref{fig:results2}b). Hence, SAM is applied on low resolution features and auxiliary task is also used at same level. For context modeling, we gather mid-level representation from multiple levels of feature extraction and merge them with SAM output. Later, a novel and efficient CAM is applied to identify relevance of each feature for label prediction.


\section{Methodology}

\begin{figure*}
    \centering
    \includegraphics[width=1\textwidth]{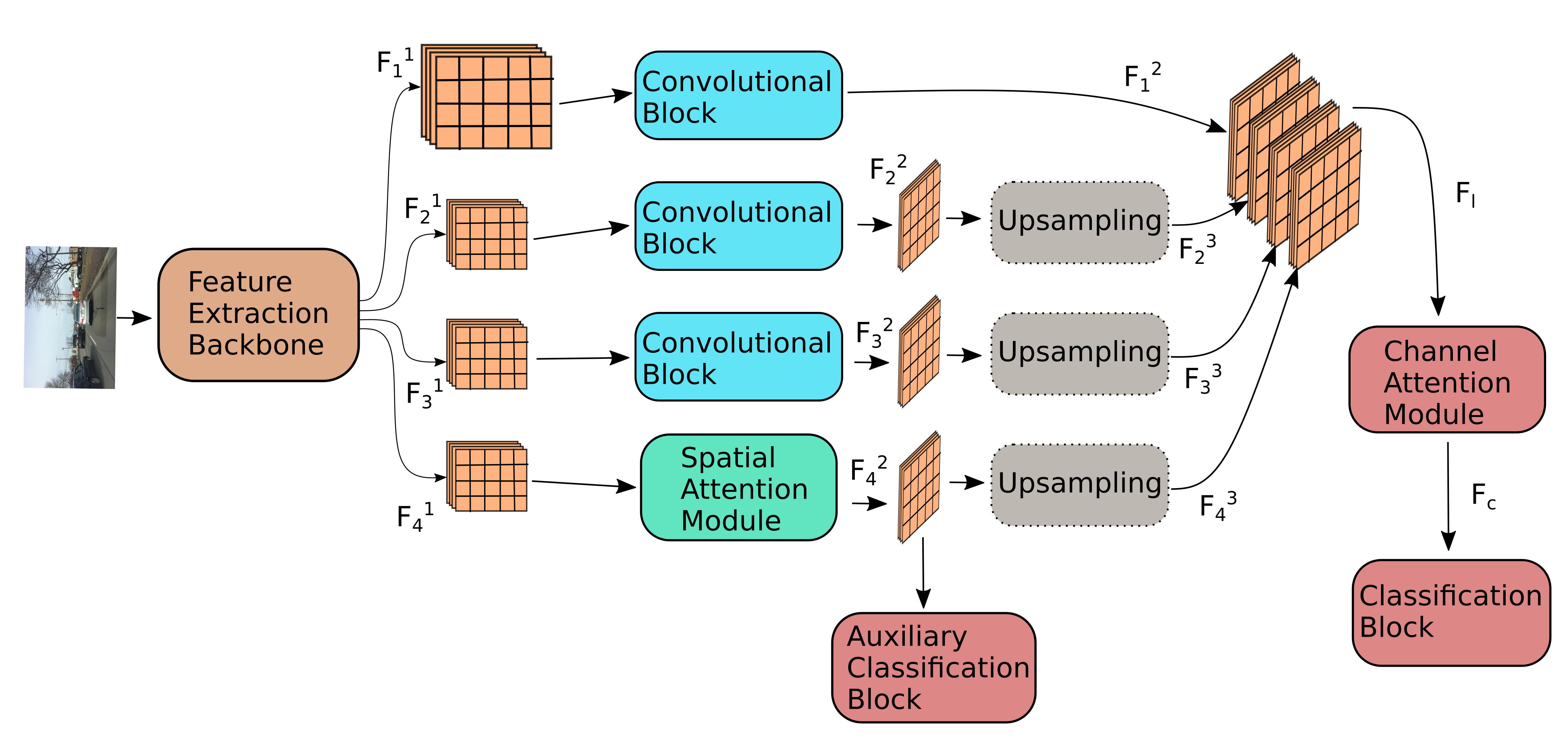}
    \caption{Complete architecture of the proposed Feature Boosting Network (FBNet).}
    \label{fig:fbnet}
\end{figure*}

In semantic scene segmentation, the goal is to predict for each pixel of an input image $I$ a label from a predefined set of classes $\{0,...,L-1\}$. All recent methods learn semantic maps at lower resolution (generally 1/8 of the input image size \cite{fu2019dual,yuan2019object,zhao2017pyramid}), whereas previous approaches used to perform semantic segmentation at the original resolution (which is generally detrimental to the quality of the output). However, when output maps have lower resolution it becomes difficult to correctly segment smaller objects/parts. To address this, our model learns semantic structures at two resolutions: 1/8th and 1/4th of the original image.

\subsection{Overall architecture}
All semantic segmentation methods employ a backbone model 
to extract good-quality feature representations. The most recent architectures use residual learning-based backbones, such as ResNet \cite{he2016deep}, ResNeSt \cite{zhang2020resnest} and ResNeXt \cite{xie2017aggregated}. The proposed architecture (Figure \ref{fig:fbnet}) uses a ResNeSt backbone due to its superior feature learning capability.

\textbf{Backbone feature extraction.} The pooling operations in last three residual blocks of ResNest are adjusted to ensure all feature maps are of the same size. Hence, feature maps produced by the first residual block are 1/4th of the original image size, and feature maps produced by the remaining three blocks are 1/8th of the original (see Fig. \ref{fig:fbnet}). Additionally, a dilation rate of 2 and 4 is used in the last two residual blocks to obtain a wider receptive field.
The final outputs of the residual blocks in the backbone are denoted by $F_1^1$, $F_2^1$, $F_3^1$ and $F_4^1$ in Fig. \ref{fig:fbnet}, respectively, in the order of their depth. The mid-level feature representations ($F_1^1$, $F_2^1$ and $F_3^1$) are transformed into more useful abstract scene-level information by applying separate convolutional blocks. 
Each of these block uses two sequential convolutional layers, each with $256$ kernels.

\textbf{Spatial attention.} Deeper features encode more abstract information about the scene. Hence, it is reasonable to hypothesize that features from the last residual block ($F_4^1$) are particularly suitable for learning spatial context.
In our model we apply a novel \emph{Spatial Attention Module} (SAM, presented in Section \ref{sec:SAM}), inspired by self-attention \cite{lin2017structured,vaswani2017attention}, to feature map $F_4^1$. The module is designed to reduce the computational cost and preserve the efficiency of the overall network. In our experiments we found that increasing the complexity of the SAM does not translate into higher model performance. Thus, in our model we compute spatial attention at lower feature resolution using a smaller attention module.
This is computationally efficient while preserving spatial invariance.

Our \textbf{feature boosting} mechanism is the composition of two operations: fusion and channel attention. 
Namely, the feature maps outputted by the last three residual blocks ($F_2^2$, $F_3^2$ and $F_4^2$) are up-sampled to match the resolution of the first (shallower) feature map ($F_1^2$). 
As these features are extracted from different depth levels of the backbone, they represent multiple scales of receptive field, with the shallower map encoding `local' feature relationships and the deeper feature maps representing `global' relationships between pixels.
All such feature maps are concatenated together along with SAM output. The first three ($F_1^2$, $F_2^3$ and $F_3^3$) have 256 channels each, while the output of the SAM ($F_4^3$) has 512 channels. A higher channel count signifies a greater contribution to the final feature vector. The final feature map ($F_l$) has 1280 channels and assembles contributions from each feature extraction stage.

\begin{figure}[h!]
    \centering
    \includegraphics[width=0.45\columnwidth]{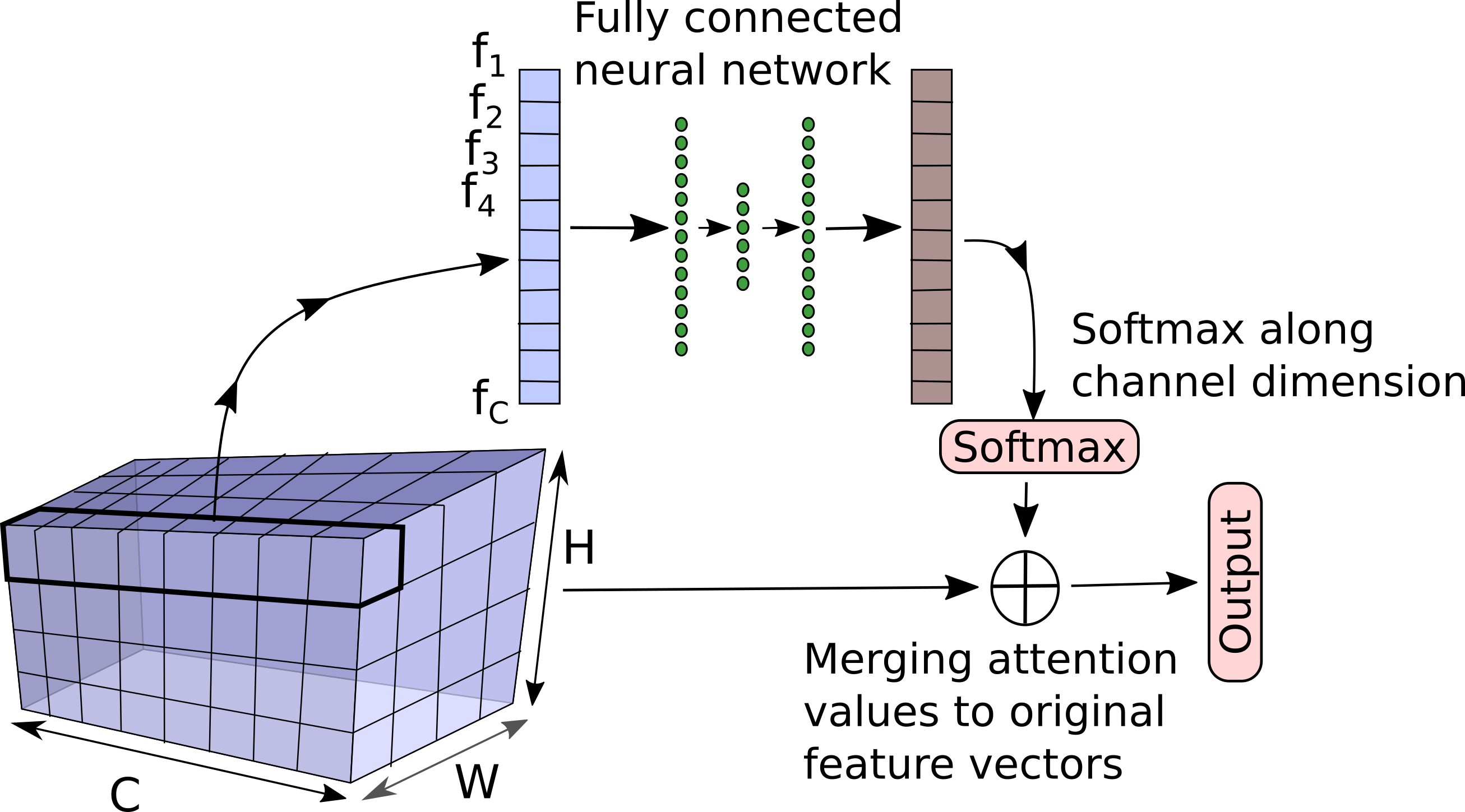}
    \caption{The proposed Channel Attention Module (CAM) used in FBNet. \vspace{-6mm}}
    \label{fig:CAM}
\end{figure}

The gathered features ($F_l$, Fig. \ref{fig:fbnet}) are forwarded to our novel \emph{Channel Attention Module} (CAM, presented in Section \ref{sec:CAM}): see Figure \ref{fig:CAM}. The latter computes the importance of each channel feature for the final prediction. As contributions from different feature extraction levels are present there, CAM helps identify what features are more effective for scale-invariant representation. The output of CAM is transferred to a classification block which uses softmax to compute the class probabilities of each pixel.

Finally, an \textbf{auxiliary task} is used to improve the feature representation in the backbone model. 
The auxiliary task is applied to the output of SAM ($F_4^2$, see bottom of Fig. \ref{fig:fbnet}). It learns same semantic segmentation targets but at half the resolution of the final output, preserving computation. 
This process helps the SAM learn spatial context in a robust fashion. The low-resolution auxiliary output also ensures that the spatial attention module focuses on the global scene context rather than overfit the local fine-grain information.

\subsection{Channel attention} \label{sec:CAM}

Let us denote the last set of feature maps after upsampling and concatenation by $y(k,i,j) \in \mathbb{R}^{C \times H \times W}$, assuming the output of the last convolutional layer has $C$ feature maps of size $H \times W$ each. In scene segmentation tasks, each feature location in each of the maps is in correspondence with a location in the image.  Hence, for each location $(i,j)$, the $C$-dimensional vector $y(\cdot, i, j)$
can be considered as the feature vector for pixel $(i,j)$ (see Figure \ref{fig:CAM}). 
The proposed CAM considers all feature locations as independent, so that pixels belonging to the same class must be represented by a similar attention vector. 

In our model, channel attention is computed by applying a fully-connected neural block composed by two layers (Fig. \ref{fig:CAM}, top).
First, pixel-specific feature vectors are compressed to develop a more robust and invariant representation. This is subsequently inflated to the same size as the original vector using a second fully connected layer. The operation is applied to each spatial location 
independently, generating an output of the same size ($\mathbb{R}^{C \times H \times W}$). A softmax activation function is used to compute the relative weight of each individual component of the vector along the $C$ dimension. This operation is performed on all locations (i.e., $H \times W$ times).
Final output of CAM is the sum of the original feature vectors and attention values.
Formally, the overall CAM transformation can be represented as:\\
\begin{equation} \label{eq:cam}
    F_c(k,i,j) = \frac{ \exp(f_c(y(k,i,j)))}{\sum_{k=1}^{C} \exp(f_c(y(k,i,j)))} + y(k,i,j),
\end{equation} \\
where $f_c$ is a function representing the fully connected layer. Softmax is computed along the channel dimension ($k$) and the result is added to the original features ($y$). This is a simple yet effective way of boosting the vector representation. 

As explained, the feature maps passed to CAM are obtained by merging feature representations computed at different stages of the architecture (see Fig. \ref{fig:fbnet}).
Hence, each component of the feature vector for a given location $(i,j)$ represents context information for the given pixel captured using different `context windows'. 

\subsection{Spatial attention} \label{sec:SAM}

The spatial attention module receives the feature maps $x(k,i,j) \in \mathbb{R}^{\Bar{C} \times \Bar{H} \times \Bar{W}}$ extracted by the final layer of the backbone, which are half the size of the final feature maps (i.e., $\Bar{H}$ and $\Bar{W}$ are 1/8 of image height and width, respectively). This, low resolution representation of the image, is designed to encode a `bigger picture' view of the scene. Here, a simple yet effective attention mechanism inspired by \emph{self-attention} \cite{lin2017structured,vaswani2017attention} is used.

First, key and query values are computed by applying a key function $f_k(.)$ and a query function $f_q(.)$ to $x(k,i,j)$. These can be represented as:

\[
s_k(k,i,j) = f_k(x(k,i,j)),
\quad
s_q(k,i,j) = f_q(x(k,i,j)),
\]
where $k$, $i$ and $j$ represent the channel, height and width dimensions of the feature map, respectively.
Both functions are computed using a 2D convolutional layer with a $1 \times 1$ kernel. The representation for both $s_k$ and $s_q$ are then transformed as in $\mathbb{R}^{\Bar{C} \times \Bar{H} \times \Bar{W}} \rightarrow \mathbb{R}^{\hat{C} \times N}$. 
Spatial attention ($S_a \in \mathbb{R}^{N \times N}$) can then be expressed as the softmax of the dot product of key and query values:
\[
    S_a = \frac{\exp(s_q^T \cdot  s_k)}{\sum_{N}\exp(s_q^T \cdot s_k)}.
\]
A value function $f_v(.)$ is then computed from the feature map $x(k,i,j)$ as in Equation \eqref{eq:value}, resulting in a mapping $\mathbb{R}^{\Bar{C} \times \Bar{H} \times \Bar{W}} \rightarrow \mathbb{R}^{\Bar{C} \times N}$, with $N=\Bar{H} \times \Bar{W}$:

\begin{equation} \label{eq:value}
    s_v(k,i,j) = f_v(x(k,i,j)).
\end{equation}
The final self-attention output is computed by taking the dot product of the attention matrix $S_a$ and the output $s_v$ of the value function, adding the initial feature map $x(k,i,j)$:

\begin{equation} \label{eq:sa_output}
    S_f = s_v^T \cdot S_a + x.
\end{equation}
Note that, before performing the addition in \eqref{eq:sa_output},
the matrix $\left[s_v^T \cdot S_a\right]$ is reshape into the shape of the input feature map $x$ via a mapping $\mathbb{R}^{\Bar{C} \times N} \rightarrow \mathbb{R}^{\Bar{C} \times \Bar{H} \times \Bar{W}}$. 

The addition of self-attention components to the original feature maps $x$ helps attenuate or amplify their respective values and is generally more effective than using only attention values or simple concatenation. The computational approach to attention proposed here is very simple and does not resort to an excessive number of parameters. In our tests we found that increasing the complexity of the spatial attention module does not significantly improve the performance of the model.

\section{Experiments}

\textbf{Datasets.} The proposed model is validated on two of the most commonly used benchmarks for scene parsing: ADE20K \cite{zhou2017scene} and Cityscapes \cite{Cordts2016Cityscapes}. ADE20K is the largest scene parsing dataset with 150 object and scene classes covering a wide range of natural scenes. The dataset contains 20,210 images for training and 2,000 images for evaluation.
Cityscapes contains images of roads, vehicles and surrounding scenes 
with a total of 2,975 training images and 500 images for validation, with for 19 semantic classes.

\textbf{Experimental setup.} For a fair and unbiased evaluation of the proposed model against the state-of-the-art, training and evaluation processes have been standardised for all models. 
We use a base size of 520 to resize the smaller side of image, and a crop size of 480 is chosen to generate a random crop from the base image. During evaluation, all images are resized to the base size. 
All models are trained for 180 epochs on the ADE20K dataset and 240 epochs on the Cityscapes dataset. For augmentation, we use random flipping and scaling in the interval $[0.5,2]$. Categorical cross entropy is used as the loss function for both the main and the auxiliary input. Stochastic Gradient Descent (SGD) is used for optimisation with momentum value of $0.9$ and weight decay value of $10^{-4}$. Initial learning rates of $0.02$ and $0.04$ are used for ADE20K and Cityscapes, respectively. Additionally, we reduce the learning rate by $1/10$ for the pre-trained backbone.
A polynomial-based learning rate reduction policy is used at training time \cite{chen2017rethinking}. The learning rate at any given iteration is obtained by scaling the base learning rate by factor of $\left(1- \frac{\text{iter}}{\text{total\_iter}}\right)^{0.9}$. 
All models are trained on eight NVIDIA RTX 6000 GPUs. Models are trained using Synchronized BN \cite{zhang2018context}. 

\textbf{Performance Metric.} Mean intersection over union (mIoU) score and pixel accuracy are used as the evaluation metrics. mIoU is the measure of overlap between the predicted object shape and ground-truth shape. Pixel accuracy is ratio of pixel correctly classified by the model to the total image pixels. Additionally, we use epoch to convergence (Epochs) to provide the insight on how fast each model achieves its best performance.

\subsection{Ablation studies} \label{sec:ablation}

\begin{table*}[h!]
    \centering
    \caption{Ablation study evaluating the performance of FBNet under different combinations of its structural components.}
    \begin{tabular}{l|c|c|c|c}
    \hline
        Model & Strategy & mIoU & Accuracy & Epochs \\
    \hline
        FBNet & SAM & 47.13 & 81.43 & 164 \\
        FBNet & CAM & 47.21 & 80.98 & 150 \\
        FBNet & FF & 47.39 & 81.16 & 156 \\
        FBNet & CAM+ SAM (parallel) & 47.55 & 81.15 & 156 \\
        FBNet & CAM+ SAM (series) & 47.95 & 81.57 & \textbf{136} \\
        FBNet & FF+ SAM+ CAM & \textbf{48.71} & \textbf{81.89} & 168 \\
    \hline
    \end{tabular}
    \label{tab:ablation1}
\end{table*}


\textbf{Significance of architectural components.} The first ablation study analyses how different components of the proposed architecture contribute to the final performance. The first two results in Table \ref{tab:ablation1} show the performance of model when only the SAM or the CAM modules are applied to the output of the backbone's final layer ($F_4^1$ in Fig. \ref{fig:fbnet}) and No feature fusion is used. The SAM model achieves a 47.13 mIoU, while CAM achieves 47.21. Hence, even without feature fusion, the CAM module is good at feature selection 
for correct prediction. Additionally, the model is faster at converging and only takes 150 epochs to reach the best mIoU.
We also tested the combination of SAM and CAM modules in series and parallel, respectively, applied in both cases only to the final layer feature maps ($F_4^1$ in Fig. \ref{fig:fbnet}). Interestingly, the series combination performs better than the individual modules as well as the parallel combination. In the series combination SAM follows CAM, whereas in parallel, the same feature map is fed to both modules and their outputs are added to generate the final attention maps.

In the last experiment, we use a feature fusion strategy where mid-level features from the backbone are combined with the output of SAM. The proposed CAM is applied to the resulting maps. The resulting model achieves significantly better performance than any of the independent modules. This further supports the hypothesis that, when applied to multi-level feature representation, CAM is able to identify the most relevant features and generate a scale invariant representation leading to much better predictions. 

\begin{table*}
    \centering
    \caption{Ablation study showing the performance of the proposed model with different state-of-the-art backbones.}
    \begin{tabular}{l|c|c|c|c}
    \hline
        Model & Backbone & MIoU & Accuracy & Epochs \\
    \hline 
        FBNet & ResNet-101 \cite{he2016deep} & 43.52 & 79.53 & 170 \\
        FBNet & ResNeXt-101 \cite{xie2017aggregated} & 43.35 & 80.09 & 180 \\
        FBNet & ResNeSt-101 \cite{zhang2020resnest} & 46.71 & 80.89 & 145 \\
        FBNet & ResNet-152 \cite{he2016deep} & 44.70 & 79.89 & \textbf{135} \\
        FBNet & ResNeSt-200 \cite{zhang2020resnest} & \textbf{48.71} & \textbf{81.89} & 168 \\
    \hline
    \end{tabular}
    \label{tab:ablation2}
\end{table*}

\textbf{Effect of different backbones.} In the second ablation study (Table \ref{tab:ablation2}), we analyse how the performance of FBNet changes when different state-of-the-art backbones are employed. The three most commonly used types of residual learning-based architectures, i.e., ResNet \cite{he2016deep}, ResNeXt \cite{xie2017aggregated} and ResNeSt \cite{zhang2020resnest}, are considered. All such backbone models are pre-trained on Imagenet and later fine-tuned as part of the proposed architecture, using a learning rate that is 1/10th of that of the whole model. 
When comparing models of similar expressive power (i.e., ResNet-101, ResNeXt-101 and ResNeSt-101), it is clear that ResNeSt-101 is the best choice for the backbone. The model achieves a 46.71 mIoU score whereas the much larger ResNet-152 only achieves 44.70 mIoU. Furthermore, ResNeSt-101 take significantly less time to train. Nevertheless, the best results are achieved when using a ResNeSt-200 backbone, with the caveat that this model takes many folds longer to train (FBNet with ResNeSt-200 takes around 90 hours).

\begin{figure}[h!]
\centering
\begin{tabular}{cc}
\includegraphics[width=5cm]{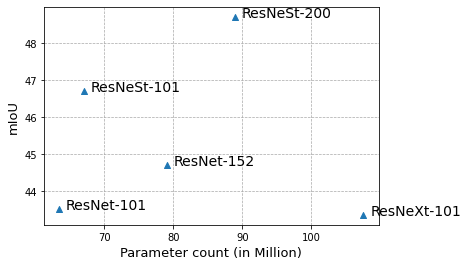}&
\includegraphics[width=4.5cm]{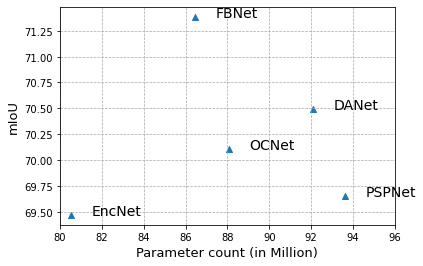}\\
(a) & (b)
\end{tabular}
    \caption{Plot of (a) mIOU achieved versus number of parameters for different backbones in Table \ref{tab:ablation2}; (b) mIoU value against number of parameters for all the models in Table \ref{tab:resnestCity}. \vspace{-4mm}}
    \label{fig:results2}
\end{figure}

\textbf{Analysis of model parameters.} Figure \ref{fig:results2}(a) plots the mIoU values achieved (for each backbone) against the number of model parameters. All models on top left side of the plot have good performance and are computational efficient (FBNet with both ResNeSt-101 and ResNeSt-200 backbones). On the other hand, models on bottom right corner have very high computational cost and lower performance (FBNet with a ResNeXt-101 backbone). Figure \ref{fig:results2}(b) shows the comparision of parameters in proposed FBNet and state-of-the-art models. All models are compared with ResNeSt-200 backbone. The proposed model only has 86.48 million parameters which is significantly lower than others. FBNet can be seen as better choice when comparing mIoU to parameter ratio.

\begin{figure*}
    \centering
    \includegraphics[width=1\textwidth]{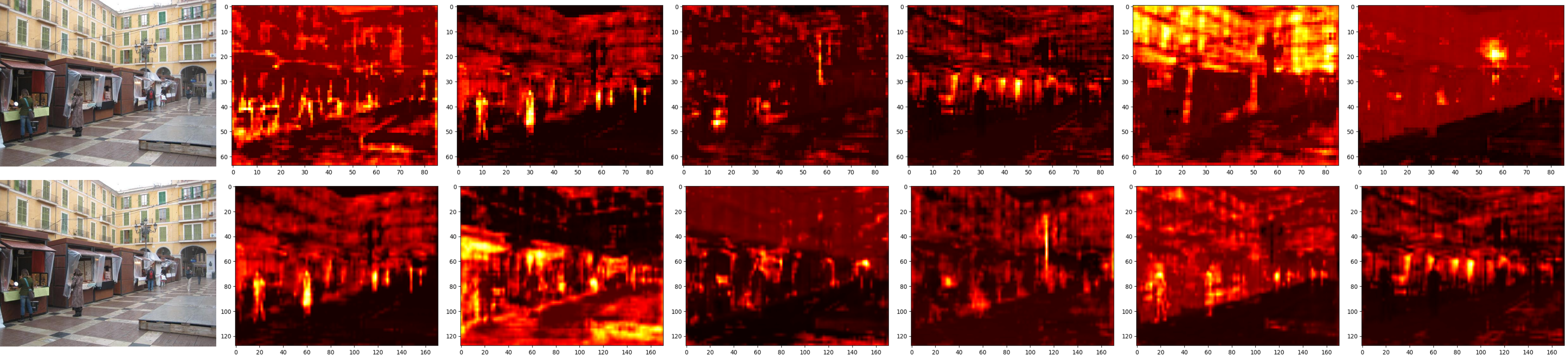}
    \vspace{-7mm}
    \caption{Attention maps for the FBNet trained on ADE20K dataset. First and second rows show attention maps for SAM (size: 64x86) and CAM (size: 128x171), respectively.}
    \label{fig:fbnet_AM}
\end{figure*}

\textbf{Attention maps for SAM and CAM .} To further elucidate the contribution of each module, we have provided attention maps for both the spatial attention module (SAM) and the channel attention module (CAM) for the model trained on the ADE20K dataset (refer Fig. \ref{fig:fbnet_AM}). The attention maps are remarkably sparse, with each map exhibiting a distinct region of focus. Certain maps concentrate on people and shops, while others highlight background structures. Notably, the spatial attention maps are half the size of the channel attention map, making the mechanism computationally highly efficient. This also enables effective translation of attention maps at higher resolutions.

\vspace{-5mm}

\vspace{-4mm}
\subsection{Comparison with the state-of-the-art}

\textbf{Experiments on ADE20K.} For unbiased evaluation we evaluated all previous state-of-the-art methods using a ResNeSt-200 backbone, even when other types of backbone were used in the original papers, as our experiments show that ResNeSt-200 \cite{zhang2020resnest} is a better backbone than ResNet-101 \cite{he2016deep}. The results for all the models are presented in Table \ref{tab:resnestAde}. The proposed FBNet model is able to achieve a mIoU of 48.71, outperforming  all the other methods. The proposed model not only achieves the best mIoU but also the best pixel accuracy, evidence of the model being able not just to predict the correct object/scene class but also to achieve a good contextual understanding of an object's neighbourhood. The second highest performance is achieved by EncNet \cite{zhang2018context}, when trained using both auxiliary and SE losses. 

\begin{figure}[!htb]
\centering
\begin{tabular}{ccccc}
\includegraphics[width=2.1cm]{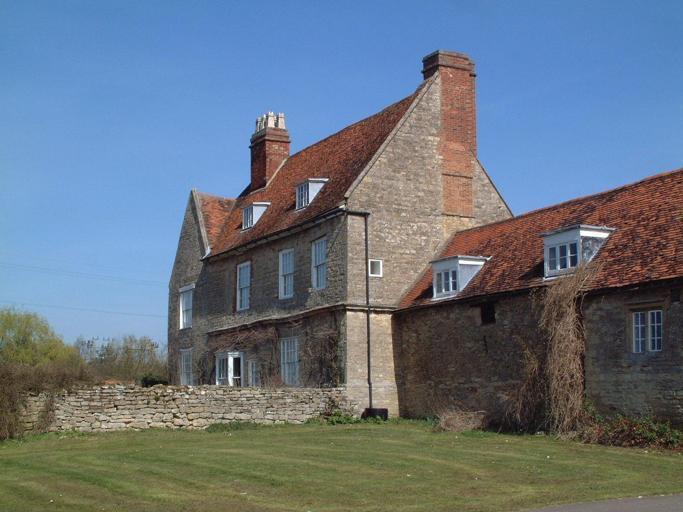}&
\includegraphics[width=2.1cm]{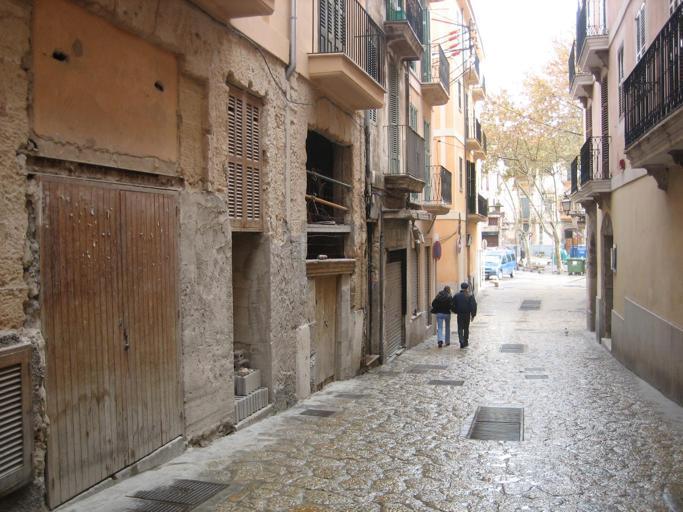}&
\includegraphics[width=2.1cm]{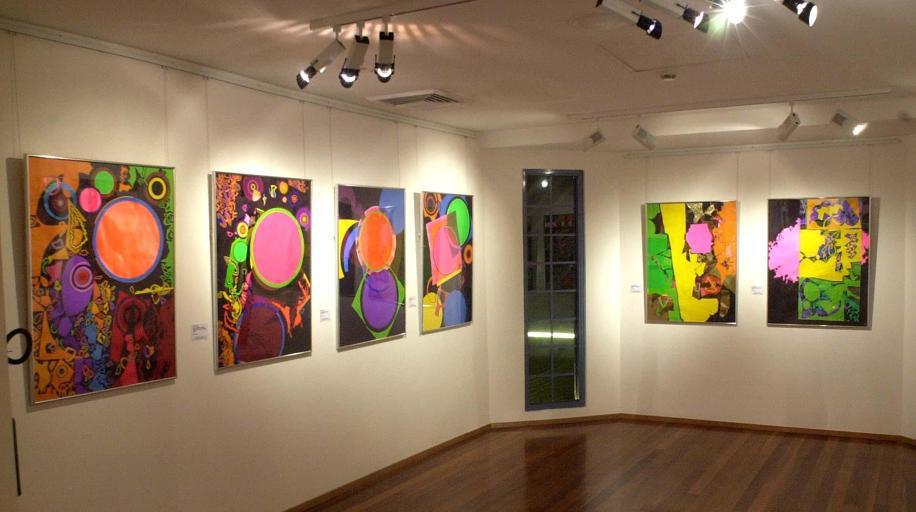}&
\includegraphics[width=2.1cm]{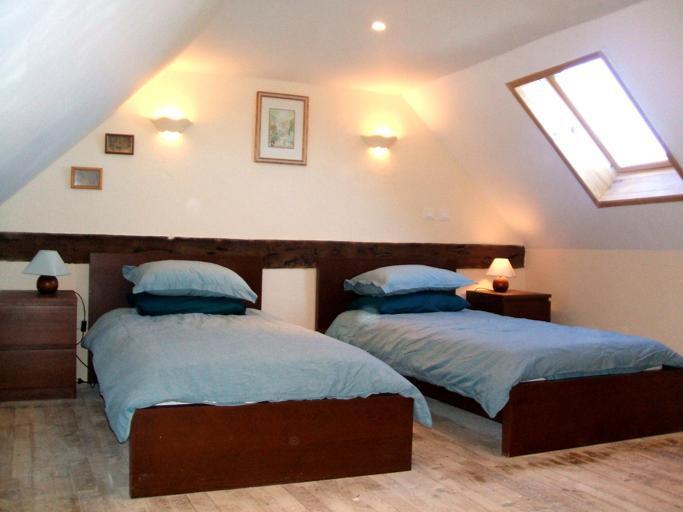}&
\includegraphics[width=2.1cm]{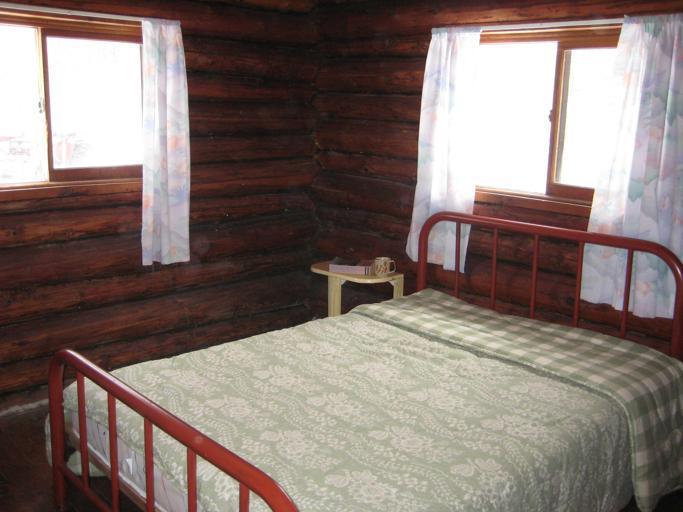}\\

\includegraphics[width=2.1cm]{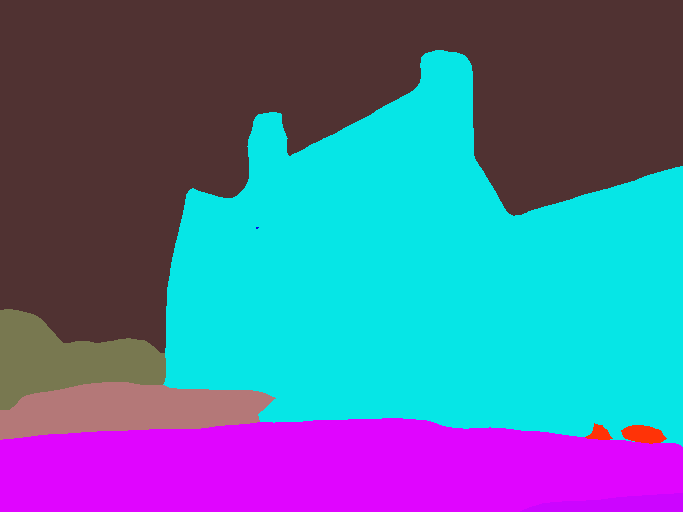}&
\includegraphics[width=2.1cm]{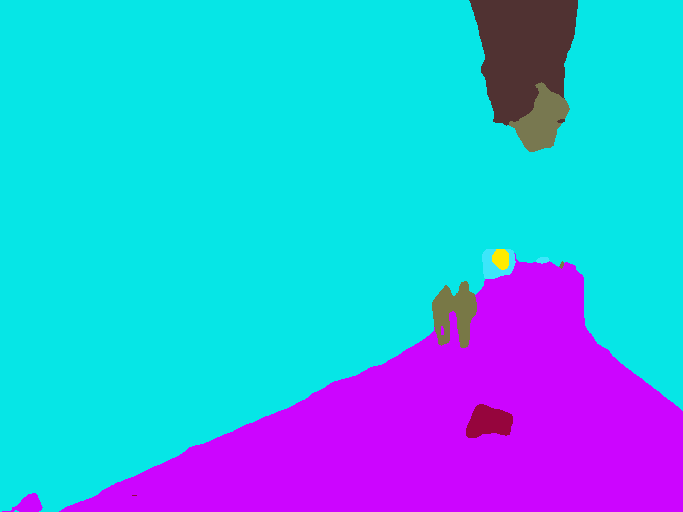}&
\includegraphics[width=2.1cm]{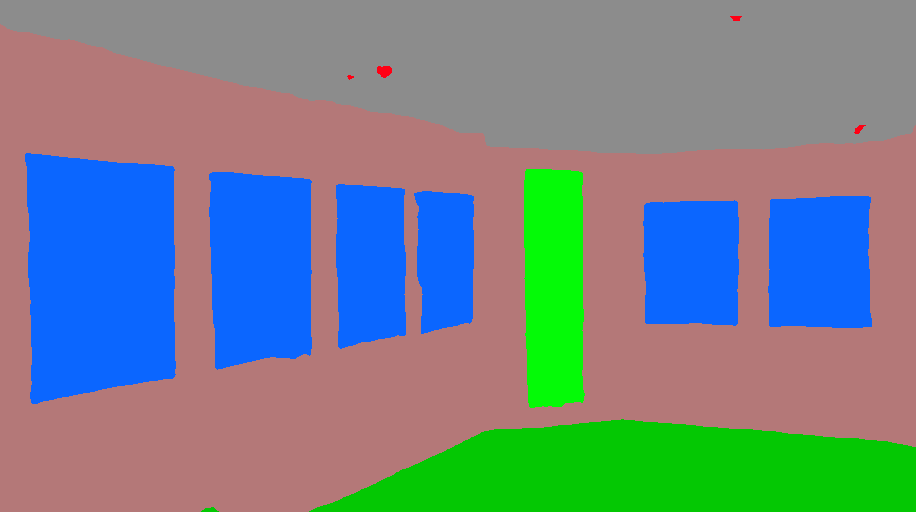}&
\includegraphics[width=2.1cm]{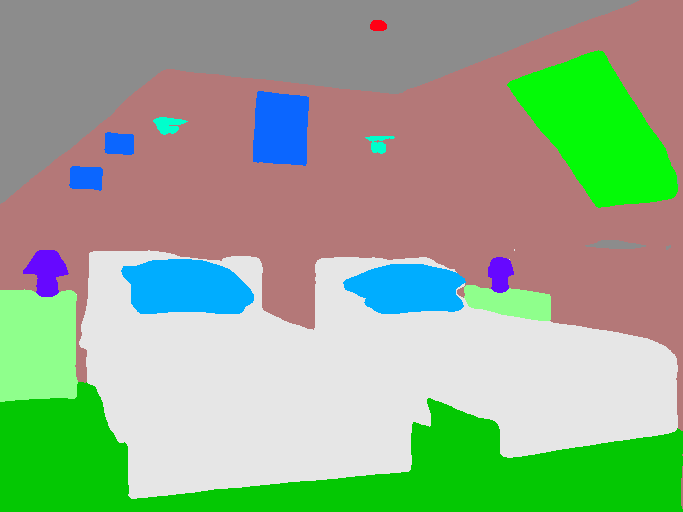}&
\includegraphics[width=2.1cm]{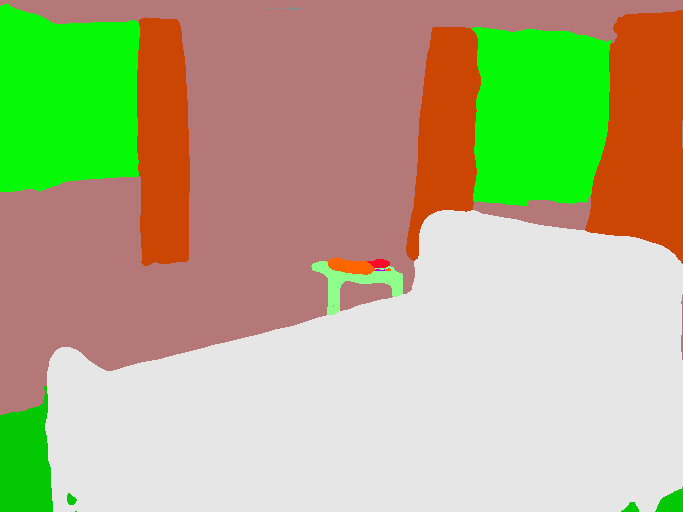}\\

\includegraphics[width=2.1cm]{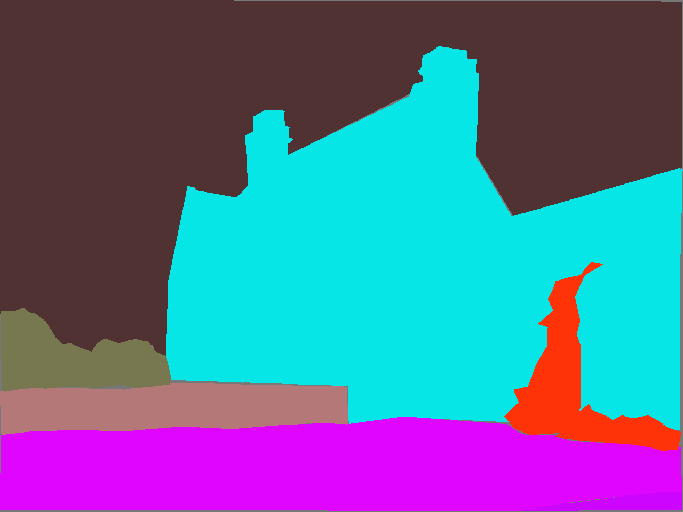}&
\includegraphics[width=2.1cm]{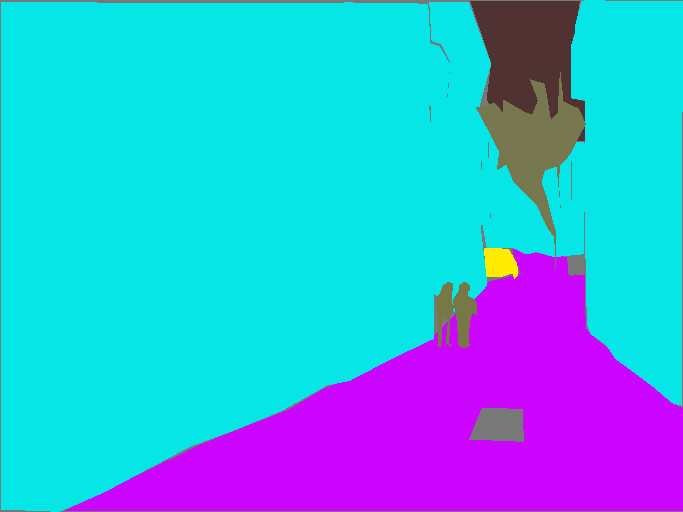}&
\includegraphics[width=2.1cm]{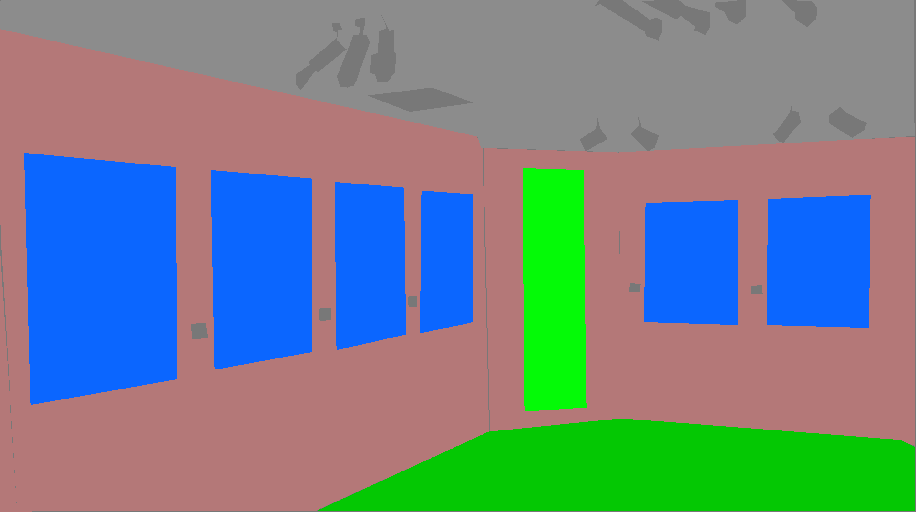}&
\includegraphics[width=2.1cm]{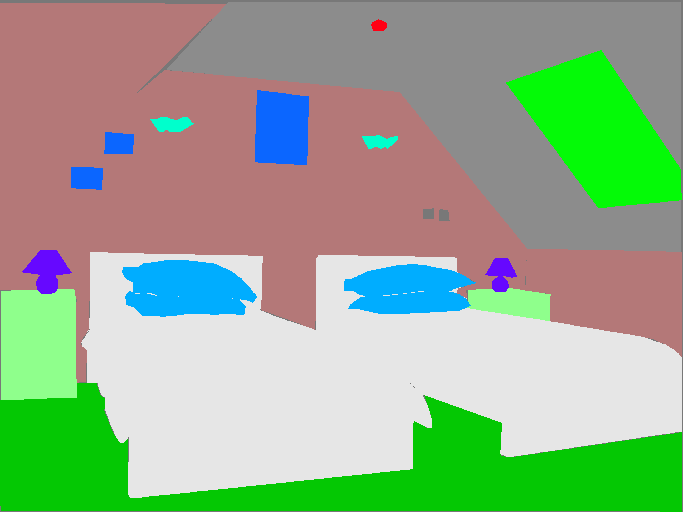}&
\includegraphics[width=2.1cm]{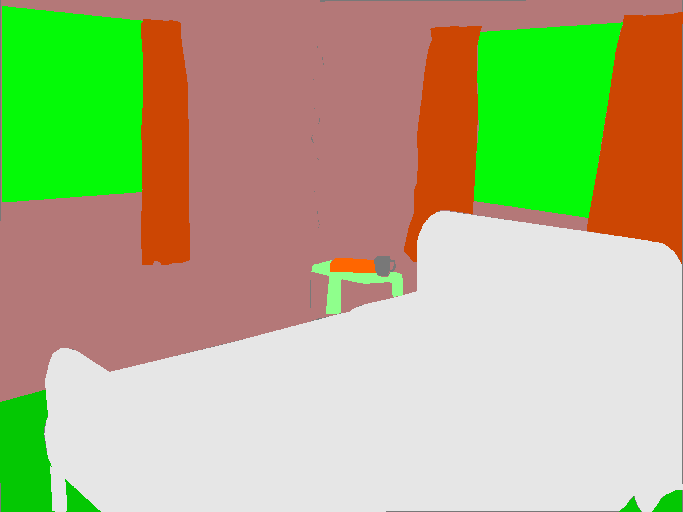}\\
\end{tabular}
    \caption{Prediction results from the proposed FBNet against the ground truth labels for a number of sample images from validation set of ADE20K dataset \cite{zhou2017scene}. The first row shows the original image, followed by the output of FBNet. The last row shows the ground truth label image.}
    \label{fig:results1}
\end{figure}

Figure \ref{fig:results1} shows some sample predictions made using FBNet against the ground truth labels. Figure contains images from both indoor and outdoor scenes with complex scene structure. All the images contain a complex scene structure where it is difficult to correctly identify pixel association in the object. The proposed model successfully predicts the scene structure that very closely resembles to the ground truth. 

\begin{table*}
    \centering
    \caption{Comparison of performance against state-of-the-art segmentation models on the ADE20k dataset.}
    \begin{tabular}{l|c|c|c|c}
        \hline
        Model & Backbone & MIoU & Accuracy & Epochs \\
        \hline
        PSPNet \cite{zhao2017pyramid} & ResNeSt-200 & 47.32 & 81.11 & 132 \\
        EncNet \cite{zhang2018context} & ResNeSt-200 & 47.80 & 81.13 & 128 \\
        DANet \cite{fu2019dual} & ResNeSt-200 & 47.75 & 80.66 & \textbf{112} \\
        OCNet \cite{yuan2018ocnet} & ResNeSt-200 & 47.38 & 81.48 & 176 \\
        FBNet (ours) & ResNeSt-200 & \textbf{48.71} & \textbf{81.89} & 168 \\
        \hline
    \end{tabular}
    \label{tab:resnestAde}
\end{table*}


\textbf{Experiments on Cityscapes.} The evaluation results on the popular Cityscapes benchmark are shown in Table \ref{tab:resnestCity}. When computing the mIoU we used the mean over the samples rather than the mean over the classes, as arguably a more expressive way of portraying true performance. 
The proposed model is again able to beat the 
previous art
by a small but significant margin. FBNet achieves an mIoU of $81.38$, whereas the second-best model (DANet) only achieves an mIoU value of $80.49$. Concerning pixel classification accuracy, again FBNet achieves the best value ($96.87$) with a significant margin from the competitors. 

\begin{table*}
    \centering
    \caption{Comparison of performance against state-of-the-art segmentation models on the Cityscapes dataset.}
    \begin{tabular}{l|c|c|c|c}
    \hline
        Model & Backbone & MIoU & Accuracy & Epochs \\
    \hline
        PSPNet \cite{zhao2017pyramid} & ResNeSt-200 & 79.65 & 96.46 & 234 \\
        EncNet \cite{zhang2018context} & ResNeSt-200 & 79.47 & 96.22 & \textbf{174} \\
        DANet \cite{fu2019dual} & ResNeSt-200 & 80.49 & 96.47 & 240 \\
        OCNet \cite{yuan2018ocnet} & ResNeSt-200 & 80.11 & 96.33 & 198 \\
        FBNet (ours) & ResNeSt-200 & \textbf{81.38} & \textbf{96.87} & 228 \\
    \hline
    \end{tabular} 
    \label{tab:resnestCity}
\end{table*}


\section{Conclusions}

In this paper we proposed a novel feature boosting network for scene parsing. The model gathers mid-level feature representations from different feature extraction stages. Subsequently a novel channel attention module is applied which computes the relevance of each feature towards computing the final class label. Additionally, a simplified self attention module is used to extract spatial contextual information before applying feature fusion. An auxiliary task is used to learn semantic features at lower resolution to achieve a better feature representation in the spatial attention module. The ablation studies are conducted structural component in the proposed model. We also studied how each backbone affects the performance of proposed model. FBNet was evaluated against other state-of-the-art models on the ADE20K and Cityscapes benchmark datasets, outperforming all the considered approaches.

\bibliography{arxiv}

\end{document}